\documentclass{isprs} % isprs class modified 23-04-2019 (Dennis Wittich)
\usepackage{subfigure}
\usepackage{setspace}
\usepackage{geometry} % added 27-02-2014 Markus Englich
\usepackage{epstopdf}
\usepackage[labelsep=period]{caption}  % added 14-04-2016 Markus Englich - Recommendation by Sebastian Brocks
\usepackage[british]{babel}%{babel} 
\usepackage[hang]{footmisc}
\usepackage{amsmath} % 导入 amsmath 宏包
\usepackage{color}

\usepackage{booktabs}
\usepackage{glossaries}
\newacronym{LoD}{LoD}{Level of Detail}
\newacronym{GML}{GML}{Geography Markup Language}
\newacronym{XML}{XML}{Extensible Markup Language}
\newacronym{OGC}{OGC}{Open Geospatial Consortium}
\newacronym{BIM}{BIM}{Building Information Modeling}
\newacronym{MLS}{MLS}{mobile laser scanning}
\newacronym{ALS}{ALS}{airborne laser scanning}
\newacronym{TLS}{TLS}{terrestrial laser scanning}
\newacronym{ICP}{ICP}{iterative closest point}
\newacronym{RANSAC}{RANSAC}{RANdom SAmple Consensus}
\newacronym{RMSE}{RMSE}{root mean squared error}
\newacronym{TIR}{TIR}{Thermal InfraRed}
\newacronym{FGR}{FGR}{Fast Global Registration}
\newacronym{GNSS}{GNSS}{global navigation satellite system}
\newacronym{IMU}{IMU}{inertial measurement units}
\newacronym{SfM}{SfM}{structure from motion}
\usepackage{todonotes}

\begin{document}

\title{Enriching Thermal Point Clouds of Buildings using Semantic 3D building Models}
\date{}

% KAO: Remove extra spacing
% Anonymous submissions, authors' names should not be visible
\author{Jingwei Zhu\textsuperscript{1}, Olaf Wysocki\textsuperscript{1}, Christoph Holst\textsuperscript{1,2}, Thomas H. Kolbe\textsuperscript{3} }%()

% KAO: Remove extra newline
% Anonymous submissions, authors' affiliations should not be visible
\address{
%TUM, Chair of Photogrammetry and Remote sensing, Chair of Engineering Geodesy, Chair of Geoinformatics 
	\textsuperscript{1 }Professorship of Photogrammetry and Remote Sensing, 
	\textsuperscript{2 }Chair of Engineering Geodesy, 
    \textsuperscript{3 }Chair of Geoinformatics, \\
 TUM School of Engineering and Design, Technical University of Munich, 80333 Munich, Germany \\ - (jingwei.zhu, olaf.wysocki, christoph.holst, thomas.kolbe)@tum.de\\

}%()

%

% If the corresponding author is NOT the final author, always add a % space before the subsequent comma, i.e.
% first author name\textsuperscript{a,}\thanks{Corresponding author} , % second author name \textsuperscript{b}, etc.
% thanks to Niclas Borlin 05-05-2016

\commission{XX, }{YY} %This field is optional. If filled, XX and YY should be replaced by adequate numbers. See https://www2.isprs.org/commissions/
\workinggroup{XX/YY} %This field is optional.
\icwg{}   %This field is optional.

% KAO: Use times symbol
\abstract{
Thermal point clouds integrate thermal radiation and laser point clouds effectively.
However, the semantic information for the interpretation of building thermal point clouds can hardly be precisely inferred.
%Semantic 3D building models at \gls{LoD}3 encapsulating facade-rich building semantics and accurate georeferencing could be the solution.
Transferring the semantics encapsulated in 3D building models at \gls{LoD}3 has a potential to fill this gap.
In this work, we propose a workflow enriching thermal point clouds with the geo-position and semantics of \gls{LoD}3 building models, which utilizes features of both modalities:
%.We also introduce a novel approach for \gls{LoD}3 model and point cloud alignment, which utilizes features of both modalities:
model point clouds are generated from \gls{LoD}3 models, and thermal point clouds are co-registered by coarse-to-fine registration.
The proposed method can automatically co-register the point clouds from different sources and enrich the thermal point cloud in facade-detailed semantics.
The enriched thermal point cloud supports thermal analysis and can facilitate the development of currently scarce deep learning models operating directly on thermal point clouds.
}

\keywords{Point clouds, LoD3 building model, Co-registration, semantic information, \gls{TIR} images}

\maketitle

%\saythanks % added 28-02-2014 Markus Englich

\section{Introduction}\label{introduction}

\sloppy

A thermal point cloud combines synchronized thermal radiation data and the laser point cloud, effectively capturing the thermal characteristics of objects.
In the case of building objects, the radiation and temperature differences can be caused by building operation, material variation, aging, and physical damage.
Therefore, thermal point clouds of buildings can be applied in hidden structure detection, energy inspection, and heritage protection~\cite{ramonReview}.
When interpreting a building's thermal point cloud, we must consider the element types and comprehensive geometric descriptions.
%\textcolor{red}{which influence the interpretation of a building's thermal point cloud.}
%The factors influencing the interpretation of a building's thermal point cloud are the element types and comprehensive geometric descriptions.
Notably, geometry-induced thermal variations such as cracks and distortions can be estimated from dense point clouds.
However, directly inferring facade elements, such as doors, windows, roofs, and walls, from thermal point clouds proves challenging.
%Geometry-related factors such as cracks and deformations can cause uneven thermal infrared distribution, and this can be supported by information obtained from dense point clouds. 

Semantic information can be obtained by manually labeling thermal point clouds or by performing semantic segmentation. 
Directly projecting \gls{TIR} images to a 3D building model for thermal point cloud generation will render unrelated objects, such as vehicles and pedestrians, to the facade, leading to misinformation. 
Therefore, it is crucial to synchronize and align the laser point cloud with \gls{TIR} images to ensure accurate radiometric representation.
However, directly labeling the laser point clouds is time-consuming and requires extensive familiarity with the study area. 
Additionally, existing methods for point cloud semantic segmenting facade elements often lack accuracy, especially for categories like clutter and windows~\cite{SU2022108372,matrone2020comparing}.
Given these challenges, it is advisable to augment the facade-level semantic information with data from reliable sources.

%Objects such as pedestrians and vehicles are recorded in the \gls{TIR} images due to the complex situation. %This synchronization .
%Existing methods for point cloud semantic segmentation for facade elements exhibit limitations: While these methods perform well for well-represented wall-like classes, achieving an accuracy of around 75\% \cite{pierdicca2020point}, and struggle with other essential classes such as doors and windows; The inference performance drops significantly for these classes scoring below 40\% due to their complex shapes and sparse representation \cite{matrone2020comparing}.

% Consequently, semantic information is vital for automatically analyzing thermal point clouds to enhance depth and reliability.
% Semantically enriched thermal building point clouds can also serve as test data, fostering the development of deep-learning-based classification and segmentation tasks where the methods and ground-truth data are scarce.
% Semantically enriched thermal building point clouds can also serve as test data fostering the development of deep-learning-based classification and segmentation tasks.
%However, the thermal intensities from single-band thermal infrared (\gls{TIR}) images are hard to interpret without materials and building element clues.

%\textcolor{red}{Semantic information about thermal point clouds should be inferred from other data resources. Accessing the original labeled point clouds is not always possible, and direct labeling the laser point cloud is time-consuming and requires familiarity with the study area.}
A potential to fill such data-scarcity gap exhibit worldwide-available semantic 3D city models.
Rich building-related semantics are encapsulated in semantic 3D building models at \gls{LoD}3, characterized by highly-detailed and object-wise semantics at the facade level \cite{grogerOGCCityGeography2012,Kolbe2021}.
Moreover, such \gls{LoD}3 models possess highly accurate absolute georeferencing accuracy, reaching up to the cm-level \cite{RoschlaubBatscheider}.
Recent trends imply growing availability of \gls{LoD}3 building models, since there are new \gls{LoD}3 datasets emerging~\footnote{https://github.com/OloOcki/awesome-citygml} as well as novel methods investigating automatic \gls{LoD}3 reconstruction \cite{wysocki2023scan2lod3,hoegner2022automatic,helmutMayerLoD3}.

Assuming the infrequent occurrence of substantial structural and morphological changes in urban architecture, we believe thermal point clouds can be semantically enriched by fusion with \gls{LoD}3 models.
However, three factors need to be considered.
% Assuming the infrequent occurrence of substantial structural and morphological changes in urban architecture, we believe thermal point clouds can be semantically enriched by fusion with recent \gls{LoD}3 models.
% However, three factors need to be considered.
First, the \gls{LoD}3 model and point clouds are organized in different data formats~\cite{Abreu2023}.
The structured \gls{LoD}3 geometric representation frequently follows the boundary representation (B-Rep), as per the CityGML standard \cite{grogerOGCCityGeography2012}. In contrast, point clouds are commonly represented as a set of unstructured points $(x, y, z)$.
Second, the shared overlap is different.
The \gls{LoD}3 model contains envelopes of the complete building, including all the outer-observable details.
The buildings in the point clouds may be incomplete due to occlusions and scanning platforms.
Besides buildings, point clouds include all the objects in the scene, such as vehicles, traffic signs, and pedestrians.
%, resulting in difficulties in registration with the  \gls{LoD}3 model
Moreover, the objects' geometric features are represented differently.
Although the \gls{LoD}3 models are highly detailed, they represent a generalized building geometry. 
In contrast, point clouds provide non-generalized, raw data representing
events and states occurring only at the time of recording, such as opened doors and blinded windows, which is also crucial for thermal analysis.

Considering all the factors, we propose a workflow to transfer the semantic information from the \gls{LoD}3 model to the thermal point clouds.
We first transfer the \gls{LoD}3 model to semantic point clouds and then co-register the point clouds.
The semantic labels are enriched in thermal point clouds according to registered building models.
The enriched thermal point clouds disclose thermal attributes for different building elements for analysis.
The co-registered point clouds, on the other hand, can help to validate the geometric information of \gls{LoD}3 models and enrich the details.
The contributions of our work include:
\begin{itemize}\label{contribution}
    \item We propose a feasible workflow to enrich the semantic information of point clouds from \gls{LoD}3 models.
    \item We propose an algorithm to automatically co-register the laser point clouds and point clouds from the LoD3 model.
    \item Our experiments validate the results of enriched information from the model and application to the thermal analysis.
\end{itemize}

The structure of this paper is organized as follows:
In Section~\ref{literature}, we summarise the related work, and our proposed methods are presented in detail in Section~\ref{method}.
Then, the data and experiments are described in Section~\ref{D&E}, and then the results are discussed in Section~\ref{result}.
Finally, some conclusions are drawn in Section~\ref{Conclusion}.

\section{Related work}\label{literature}

\subsection{Semantic 3D building models}

Semantic 3D city models comprehensively describe structures, taxonomies, and aggregations on a city, regional, and even national scale. 
Internationally, the standard CityGML, established by the \gls{OGC} \cite{Kolbe2009,grogerOGCCityGeography2012,kolbeOGCCityGeography2021}, is utilized for the representation and management of city models. 
CityGML facilitates the modeling of urban objects with their 3D geometry, appearance, topology, and semantics at four different \gls{LoD}. 
The latest data model of CityGML 3.0 adheres to the ISO 191xx series of geographic information standards, and CityGML datasets are commonly encoded using either the \gls{GML} or CityJSON \cite{Kutzner2020,ledoux2019cityjson}.

Since urban dwellings are the cornerstone of each city, most existing semantic 3D city models comprise buildings \cite{biljeckiApplications3DCity2015}.
\gls{LoD}1 and \gls{LoD}2 building models are currently widely available, as underscored by the example of approximately 220 million models available in Germany, Japan, the Netherlands, Switzerland, the United States, and Poland~\footnote{https://github.com/OloOcki/awesome-citygml}.
This broad adoption owes mainly to the robust 3D reconstruction algorithms and available building footprints combined with aerial observations \cite{RoschlaubBatscheider,HAALA2010570}.
Although \gls{LoD}1 and \gls{LoD}2 possess building semantics, they lack detailed facade semantics, which is pivotal for facade-level point cloud labeling. 
This gap can be filled by \gls{LoD}3 building models, characterized by descriptive facades, composed of objects such as windows, doors, balconies, and even underpasses \cite{wysockiUnderpasses}. 
Currently, the automatic \gls{LoD}3 reconstruction is an active field of research proposing various promising methods and input datasets to solve the challenge \cite{wysocki2023scan2lod3,hoegner2022automatic,helmutMayerLoD3}.

\subsection{Model to Point Clouds Registration}
Registration of 3D models and point clouds is typically done by feature matching.
Point clouds offer accurate and detailed geometry information about existing structures. 
Therefore, they are widely used as data providers for 3D model reconstruction and manual modeling.
Extracted planes are used as features for co-registration due to their relatively simple representation and frequent occurrence in man-made objects.
\cite{Bosche2012} propose the semi-automatic method to register construction sites with \gls{BIM} models. Three corresponding planes are required to be manually selected for coarse registration. 
\cite{Gruner2022} also focus on planes but generalize \gls{BIM} faces as \gls{TLS} patches with a point and a normal vector.
The detected faces and planes are organized manually by connected relation.
Then, the model faces are co-registered to the point cloud patches for monitoring the construction process. 
Besides man-involved work, automatic methods are also investigated.
\cite{Sheik2022} group the detected parallel planes as descriptors to register as-built point clouds and as-plan \gls{BIM} models.
The use of planes avoids setting control points for registration, but a sufficient number and unique patterns are required.
Another often-used set of primitives is lines.
\cite{Kaiser2022} propose a fully automated method to register photogrammetric point clouds to a building model with lines for the indoor scene, where sufficient and well-distributed corresponding line features from images are required.
\cite{Chen2022} conducts a coarse-to-fine registration from \gls{BIM} to the point cloud.
Raw camera poses are used to coarsely align the model, and then an adopted \gls{ICP} to achieve the fine registration.

Despite registering point clouds to \gls{BIM} models, which targets to monitor and detect the changes, only a limited number of research groups have delved into the realm of coregistration between street-level point clouds and semantic 3D city models in our extensive investigation.%~\cite{goebbels2018line, goebbels2018linear, goebbels2019iterative, lucks2021improving}. 
%Among these, the works of Goebbels et al.~\cite{goebbels2018line, goebbels2018linear, goebbels2019iterative} stand out prominently. 
\cite{goebbels2019iterative} primarily centers around point clouds generated from images through the \gls{SfM} algorithm, utilizing radiometric features for prefiltering, which may inadvertently eliminate valid building features.
\cite{goebbels2018line} detect the footprint of buildings from point clouds and \gls{LoD} model. A mixed integer linear program is employed to identify correspondences between 2D lines and points.
In these cases, sufficient line and point features are necessary to form unique patterns for registration.
\cite{lucks2021improving} consider only façades points by random forest and register to \gls{LoD}1 model for trajectory optimization.
This approach, however, requires training data and initial transformation information.
%In one of the works Goebbels et al.~\cite{goebbels2018line, goebbels2018linear} employ a Mixed Integer Linear Program to identify correspondences across modalities. 
%It is noteworthy that Goebbels et al.'s publications primarily center around point clouds generated from images through the \gls{SFM} algorithm, necessitating the utilization of radiometric features for prefiltering~\cite{goebbels2019iterative}, which may inadvertently eliminate valid building features.
%In a distinct approach, Lucks~\cite{lucks2021improving} incorporates the ICP point-to-plane algorithm for matching \gls{MLS} point clouds with semantic 3D city models. To enhance matching accuracy, Lucks et al. introduce a random forest methodology to selectively consider only those points in point clouds that depict façades.

%However, only a limited number of research groups have delved into the realm of coregistration between point clouds and semantic 3D city models based on our extensive investigation~\cite{goebbels2018line,goebbels2018linear,goebbels2019iterative,lucks2021improving}. 
%In most cases, a coarse pos is required for registration.

\subsection{Point cloud co-registration}
Point cloud registration has long been a research topic.
It involves aligning point clouds from different sources, with low overlap, or from metrically inaccurate datasets. 
The standard registration, such as \gls{ICP}~\cite{segal2009generalized}, can deal with common situations with sufficient overlap and similar point density but can hardly handle complex situations.
Using control points~(targets) could solve this issue leading to a high and traceable accuracy~\cite{JanßenKuhlmannHolst+2022+91+106,JanßenKuhlmannHolst+2023}, but requires manual field work and is thus not scalable to large data sets and it lacks autonomy.
Therefore, most point cloud registration pipelines inherently have a workflow containing two steps: coarse registration for initial transformation and refined transformation with denser correspondences~\cite{xu2023point}.
The coarse transformation uses sparse feature-based correspondences, and it is crucial for non-georeferenced point clouds.
Key points~\cite{barnea2008keypoint}, lines~\cite{chen2019feature}, and planes~\cite{li2022point} can all be used as features for registration.
Moreover, a combination of the features is also used, like 4PCS~\cite{aiger20084} and Super4PCS~\cite{mellado2014super}.
When the initial relative poses are given by \gls{GNSS} or manually processed, the coarse registration is typically unnecessary.
The fine registration can adjust the geometric transformation to achieve better accuracy.
It usually iteratively updates the transformation matrix to minimize the point distances with denser correspondences.
The \gls{ICP} method is widely adopted in the field owing to its simplicity and efficiency. 
Numerous algorithms have stemmed from the \gls{ICP} framework, exemplified by references such as \cite{yang2015go}.
% In contrast to traditional methods, deep learning has become increasingly prevalent for point cloud registration, as indicated by literature such as \cite{lu2019deepvcp}.
Recently, deep learning~\cite{lu2019deepvcp,rs14122883} methods have become increasingly prevalent for point cloud registration.
However, a notable challenge persists in managing the intricacies of large study areas and meeting the demand for extensive training datasets.
This issue underscores the current limitations in the application of deep learning techniques to point cloud registration, particularly in addressing the complexities posed by expansive geographical contexts.

 \section{Method}\label{method}

\begin{figure}[h!]
     \centering
     \includegraphics[width=1.0 \columnwidth]{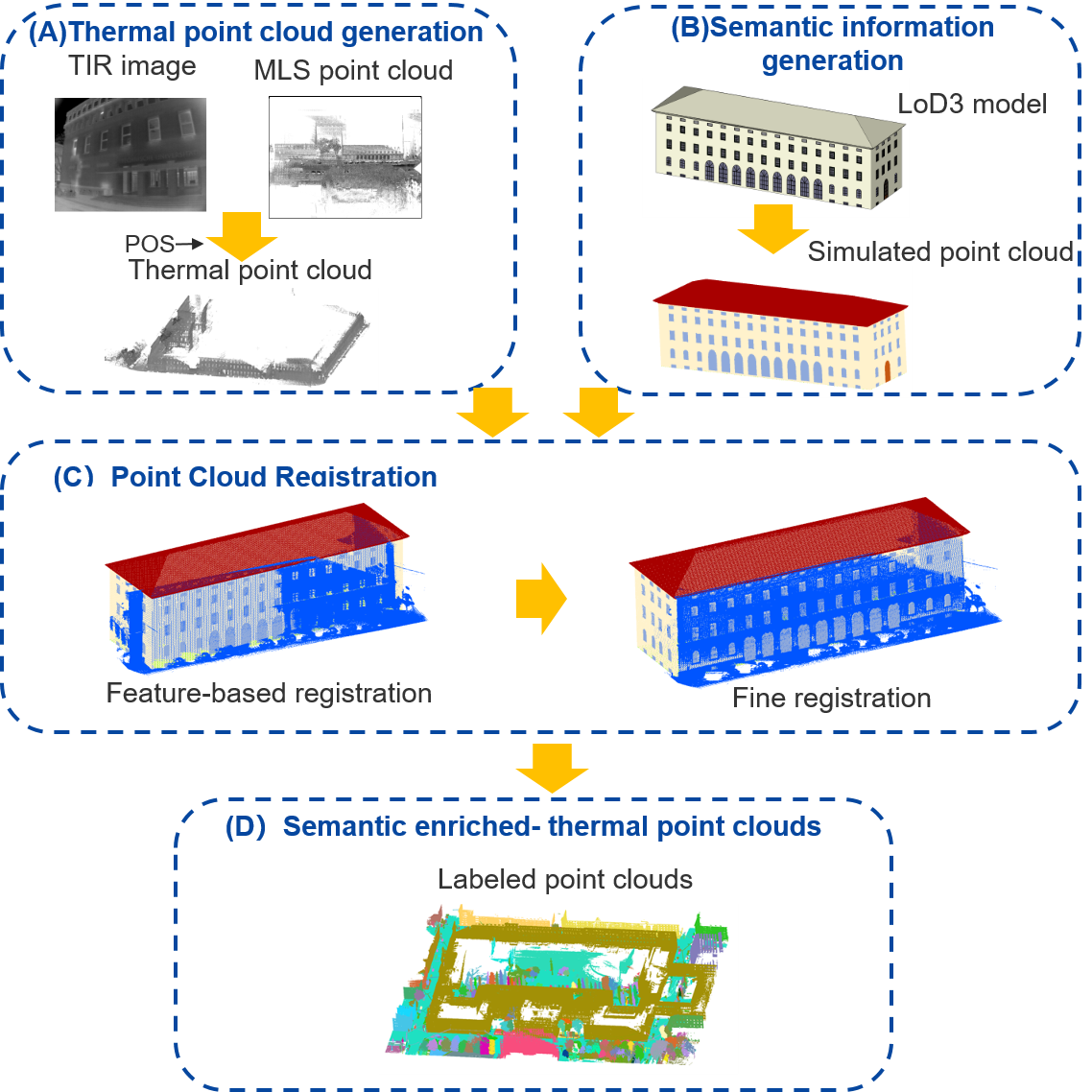}
     \caption{The general workflow}
     \label{fig:workflow}
 \end{figure}

Our work aims to transfer the semantic information from the \gls{LoD}3 model to the corresponding thermal point clouds.
The general workflow is shown in Figure~\ref{fig:workflow}. 
First, the point clouds are generated from laser point clouds and \gls{TIR} image sequence, and an \gls{LoD}3 model respectively.
Then, thermal point clouds are aligned to the model point clouds to obtain the transformation matrix.
After registration, the semantic information from model point clouds is transferred to the thermal point clouds for analysis.

\subsection{Point cloud generation}\label{point_clouds}

% To cope with the cross-domain gap, the \gls{LoD}3 model is first converted to a point cloud.
% Compared to the comprehensive information provided by the well-structured \gls{LoD}3 model, the point cloud may exhibit information gaps due to occlusions or limitations in scanning modes.
% The planes detected from point clouds may not always correspond to those in the model. 
% Therefore, features in windows and doors are needed for matching. 
% Consequently, the envelope of the \gls{LoD}3 model is resampled to the point cloud.
To cope with the cross-domain gap, we opt to homogenize the two distinct representations. 
The approach of generalizing the point cloud feature to match the model primitives is challenging, as shown by the 3D building reconstruction research~\cite{wysocki2023scan2lod3}.
The occlusion and incompleteness of laser point clouds lead to inefficiency and false matching.
Therefore, \gls{LoD}3 geometry is sub-sampled to a set of points corresponding to the representation of the thermal point cloud.
%Owing to a low-level and homogeneous representation, we operate on two sets of points and leverage advancements in point-to-point co-registration methods. 

%Considering the data differences, the \gls{LoD}3 model and point clouds should be converted to the same format.
%\gls{LoD}3 building models describe the building in a highly abstract structure by elements, while point clouds depict the as-built representation of the same target.
%Compared to segmenting and organizing the raw point clouds into structured representations, sampling the reconstructed model into point clouds is more rational.
%We opt for the model sampling, as it is feasible to  resample the structured information into point set.

\textbf{Thermal point cloud generation}
The thermal point clouds are generated by projecting the thermal texture from \gls{TIR} images onto the \gls{MLS} point clouds with position information \cite{zhu2023generation}.
The \gls{MLS} point clouds and \gls{TIR} image sequences are captured with the same platform. 
%The point clouds simulate the representation of the test areas with similar shapes and features.
When the relative poses of the thermal camera are estimated, the points in the point clouds find its corresponding points in the \gls{TIR} image by co-lineary equation~(eq.~\ref{eq:projection})~\cite{zhu2021fusion}.
%The orientation of the sensors is roughly known from a GPS/INS unit and measured lever arm.

\begin{equation}\label{eq:projection}
	u_i=K[R|T]X_i
\end{equation}
\noindent 

\begin{tabbing} 
where \hspace{0.6cm} \= $u_i$ = image coordinates\\
\> $X_i$ = point cloud coordinates\\
\> $K$ = camera parameters, including aspect ratio $s$,\\ \hspace{2.0cm} focal length $f$ and principle points $(c_x, c_y)$\\
\> $R$ = $3\times 3$ rotation matrix\\
\> $T$ = translation matrix
\end{tabbing}

\begin{equation}\label{eq:ex-in}
    K=
    \begin{bmatrix}
    s\cdot f & 0 & c_x\\
    0&f & c_y\\
    0& 0 & 1
    \end{bmatrix}
     R=\begin{bmatrix}
        \alpha_{11},\alpha_{12},\alpha_{13}\\
        \alpha_{21},\alpha_{22},\alpha_{23}\\
        \alpha_{31},\alpha_{32},\alpha_{33}\\
    \end{bmatrix},T=\begin{bmatrix}
        t_1\\t_2\\t_3
    \end{bmatrix}
\end{equation}
When the matrix $K$, $R$ and $T$ are obtained from pose parameters, the corresponding intensity values of point clouds are calculated from the image and rendered to the point cloud.

\textbf{Model point cloud generation}
The semantic 3D building models follow the paradigm of boundary representation (B-Rep), where each modeled object has its geometrical, outer-observable surface explicitly described by a set of vertices \cite{Kolbe2021}.
Moreover, each object in the model has assigned semantics and shall not overlap with other objects.
\begin{figure}[h!]
     \centering
     \includegraphics[width=0.7 \columnwidth]{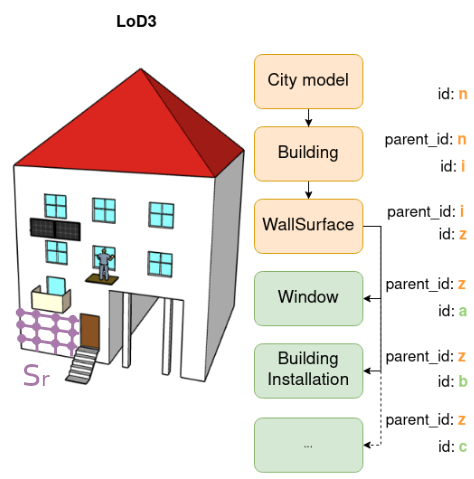}
     \caption{Semantic \gls{LoD}3 building model hierarchy comprises LoD2 (orange boxes) and \gls{LoD}3 (green boxes) features, where each object has a unique identifier ($id$) and class ($label$). Our object-level sampling approach is shown on a WallSurface in purple, where lines indicate the distance ($s_{r}$) between surface-sampled points. Adapted from \protect\cite{wysockiMLS2LoD3}}
     \label{fig:lod3hierarchy}
 \end{figure}
As illustrated in Figure~\ref{fig:lod3hierarchy}, we leverage these traits to our advantage by performing object-oriented point surface sampling on a regular grid (purple) at the given sampling rate $s_{r}$; the parameter is chosen accordingly to the expected thermal point cloud density. 
Each sampled point inherits the semantic class of a leaf object $label_{i}$ and its associated unique identifier $id_{i}$.
This results in a point cloud $MC_i$ extended by a scalar value $label_{i}$ and $id_{i}$, as in Equation \ref{eq:pointCloudModel}.
\begin{equation} \label{eq:pointCloudModel}
    MC_i = [x_{i}, y_{i}, z_{i}, label_{i}, id_{i}]
\end{equation}

 \subsection{Co-registration}\label{registration}

% After point cloud generation, turned of point clouds and \gls{LoD}3 model  to point cloud registration.
Owing to the point cloud generation step, we formulate the alignment as the cloud-to-cloud co-registration problem.
In our work, only rigid transformations between point clouds are considered, as shown in eq.~\ref{rigid}.

\begin{equation}\label{rigid}
    MC_i=[R_p|T_p]TC_i
\end{equation}

\begin{tabbing}
    where \hspace{0.6cm} \= $TC_i$ = coordinates of thermal point clouds\\
    \> $MC_i$ = points in the model point clouds
\end{tabbing}

Considering the limited overlapping areas, the features and details of building facades are important for co-registration with point clouds.
Unlike \gls{LoD}1 and \gls{LoD}2, the \gls{LoD}3 building models comprise 3D facade elements \cite{Kolbe2021}.
% The \gls{LoD}3 models have advantages of more information about windows and doors compared to LoD1 and LoD2.
The increased information makes it possible to locate the corresponding features with point cloud facades, especially in highly repeated patterns.
However, the features of the windows and doors may not have the same representation in measured point cloud and the model point cloud of the \gls{LoD}3 model. 
%The thermal point clouds contain all the status differences and occlusions.
One-to-one corresponding features cannot be guaranteed. 
Under this condition, we propose a coarse-to-fine registration by combining a feature-based method using \gls{FGR}~\cite{zhou2016fast} and our adopted version of plane-based \gls{ICP}~\cite{Rusinkiewicz2001,wysocki2021unlocking} as fine registration.

\gls{FGR} uses features to calculate the correspondences and estimate the transformation matrix.
It first calculates the FPFH~(Fast Point Feature Histograms) features of the point clouds.
The initial corresponding points are established by feature matching with nearest neighbor pairs. 
It uses the Reciprocity test and Tuple test to improve the inlier ratio of the correspondences set.
Due to the differences in feature representation, noisy correspondences cannot be avoided.
Then, the pose are optimized such that distances between corresponding points are minimized.
The optimization function for the optimal transformation matrix estimation is expressed as in~(eq.~\ref{optimization}):

\begin{equation}\label{optimization}
    E([R|T]) = \arg\min_{} \sum_{(p,q)\in K} \rho\lVert p_i - [R|T] \cdot q_i \rVert
\end{equation}
\begin{tabbing}
    where \hspace{0.6cm} \= $p_i$ = point coordinates in thermal point clouds\\
    \> $q_i$ = corresponding point coordinates \\
    \hspace{2.0cm} in the model point clouds\\
    \> $\rho$ = robust penalty
\end{tabbing}

%The optimization function can be summarized as eq.~\ref{optimization}
\gls{FGR} uses a scaled, well-chosen German-McClure estimator to reduce the computation, and Black-Rangarajan duality is used to optimize eq.~\ref{optimization} with a line process over the correspondences.
Then, the optimization objective can be turned into a least-squares objective and the Gauss-Newton method is used to find the solution.

After coarse registration, a fine registration is further updated by point-to-plane \gls{ICP}.
Although the \gls{FGR} provides the initial transformation result, it is insufficient for detailed analysis due to the false matching from different target and source point cloud feature representation.
%Therefore, the fine registration is required.
We adopt the point-to-plane \gls{ICP} variant \cite{Rusinkiewicz2001} and model-based height rectification~\cite{wysocki2021unlocking} with both height and center point rectification.
Assuming that the model and point clouds are all related to the ground, the thermal point cloud is lifted to the same basic height of the model.
The same applies for the center points of the planes.
We leverage the algorithm to align two point clouds while minimizing the distances between corresponding points belonging to the target and source plane.
Since thermal point clouds only capture the facades of buildings, which follow planar-like shapes, the main plane of the buildings are extracted as a base for the registration.
We perform the plane extraction using \gls{RANSAC} algorithm \cite{schnabel2007efficient} where the main planes are detected in both point clouds. 
Our approach assumes that the closest planes are detected and aligned from the coarse registration.
The algorithm initiates with an initial alignment represented as identity matrix, where the target point cloud is approximately aligned with the source point cloud using an initial transformation matrix of from coarse registration. Subsequently, a nearest neighbor search is conducted for each point in the source cloud to find its closest counterpart in the target cloud with optimization eq.~\ref{eq:plane_icp} 

\begin{equation}\label{eq:plane_icp}
    E([R|T]) = \arg\min \sum_{(p,q)\in K} \lVert p_i - [R|T] \cdot q_i \rVert n_p
\end{equation}

\begin{tabbing}
    where \hspace{0.6cm} \= $n_p$ = normal of the point $p$
\end{tabbing}

The maximal corresponding distance corresponds to the $d_{max}$.
The convergence criteria are met if the \gls{RMSE} reaches $t_{rmse}$~(eq.\ref{eq:rmse}) threshold and performs $t_{it}$ iterations.

\begin{equation}\label{eq:rmse}
 t_{rmse} = \sqrt{\frac{1}{N} \sum_{i=1}^{N} (d_{i})^2}
\end{equation}

Upon meeting these conditions, the algorithm concludes, providing the final alignment result in the form of a transformation matrix. 
The updated transformation parameters are used to refine the position of the source point cloud. 
% Since the traditional point-to-plane method may induce height misalignment, we rectify the height based on the model point cloud minimum height.
% We calculate the Euclidean distance between the lowest points of both target and source point cloud and apply the difference to the source point cloud.

 \subsection{Semantic enrichment}\label{enrichment}

 In Section~\ref{registration}, the transformation matrix is calculated to register the thermal point clouds to the model point clouds.
 The geo-reference coordinates of thermal point clouds can be calculated by applying the estimated transformation matrix.
 After the transformation, the thermal point cloud is aligned to the model point cloud.
 Assuming there are no changes in building details for the laser point clouds and model; the semantic labels for windows and doors will remain the same.
 Therefore, the points in thermal point clouds should have the same labels as in the model point clouds.
 Considering the differences in sampling rate and locations, a threshold distance is set to minimize the false correspondence.
 For each point in the thermal point clouds, the closest point in the model point clouds is calculated, and the label is given to the laser point.
 If the nearest neighbor points do not have corresponding labeled points in the model point cloud within a certain threshold, they can be regarded as noise and labeled as "unlabeled".
 This avoids mismatched labels from other objects, such as trees and pedestrians, while keeping the labels for the buildings.

 \section{Data and Experiments}\label{D&E}
\begin{figure}[h!]
    \centering
    \includegraphics[width=0.8\columnwidth]{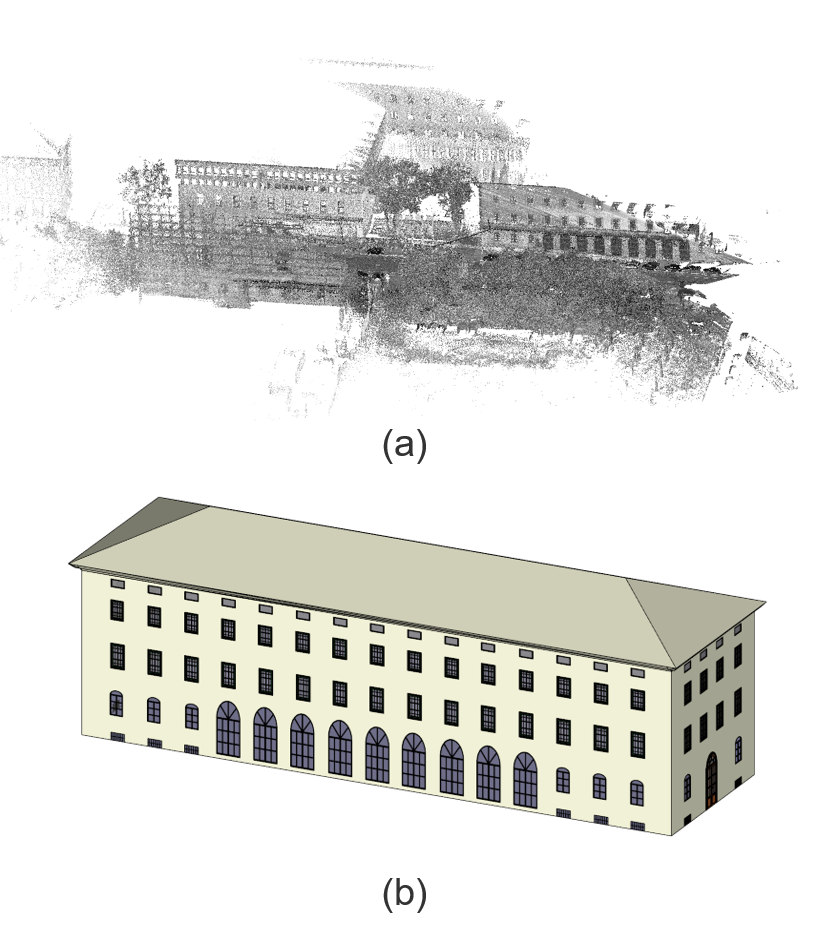}
    \caption{(a) Original MLS point cloud and (b) \gls{LoD}3 model}
    \label{fig:data}
\end{figure}

The test site is around the main campus of the Techinical University of Munich~(TUM). 
The thermal point clouds were generated by combining thermal image sequences and MLS point clouds from the TUM-MLS dataset~\cite{zhu2020}.%TUM-MLS,MODISSA
The TUM-MLS dataset was measured using a mobile platform, which includes two laser scanners and a thermal camera.
The poses of \gls{TIR} images were estimated from the \gls{GNSS} and the \gls{IMU} system from the integrated platform.
The \gls{LoD}3 model was selected from the TUM2TWIN dataset~~\footnote{https://github.com/tum-gis/tum2twin}.
We selected one building from the test dataset close to the main gate characterized by a partial coverage (the so-called \textit{building 23}), as shown in Figure~\ref{fig:data}.
We set the sampling rate at $s_{r} = 0.1$, which resulted in a uniformly sampled point cloud of a 0.1 m distance.
Figure.~\ref{fig:data}(a) shows an example of the TUM-MLS point cloud with intensity, and (b) demonstrates the \gls{LoD}3 building model.

The generation of thermal point clouds and semantic labeling was done using c++ and pcl library(1.81) \cite{rusu20113d}.
With 32G RAM, and an i7-6000 @3.4 GHz CPU, it takes approximately 346.06s for labeling.
The \gls{FGR} was performed using the code from~\cite{zhou2016fast}.
Further experiments were performed using the Feature Manipulation Engine (FME)  version 2020.01 and Open3D \cite{zhou2018open3d}.
The implementation is available in a public repository\footnote{https://github.com/tum-pf/LoD3toTCld}.

 \section{Result and Discussion}\label{result}

\begin{figure}[h!]
    \centering
    \includegraphics[width=0.8\columnwidth]{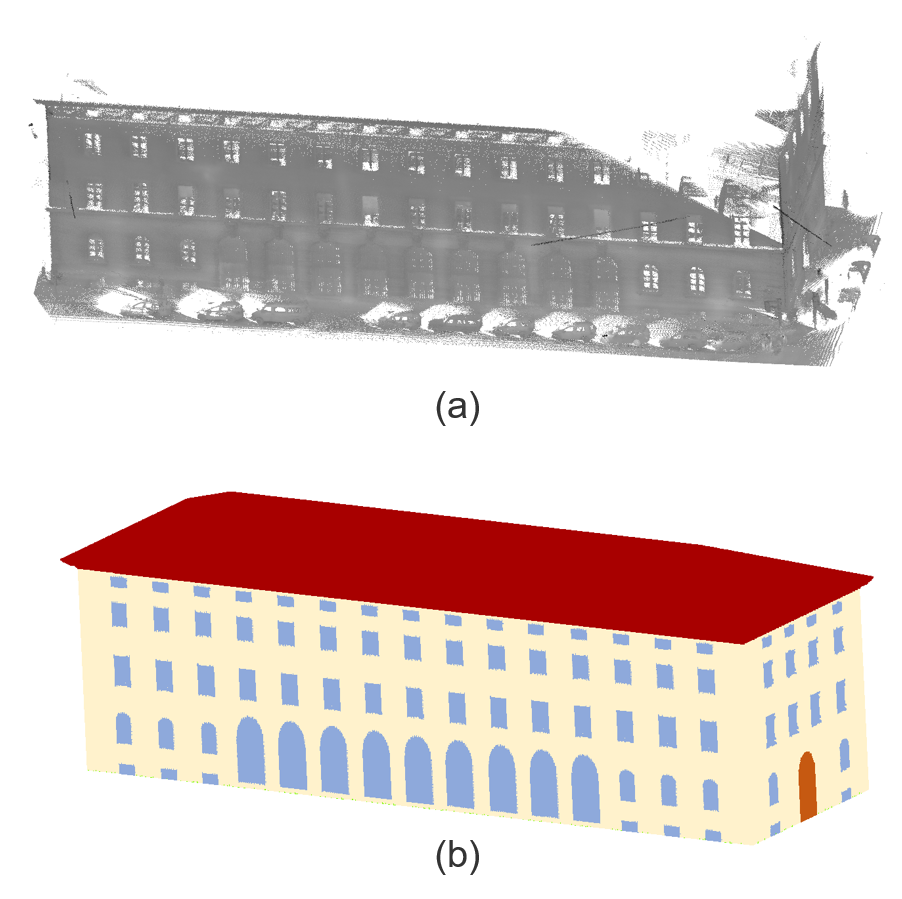}
    \caption{Generated point clouds. (a) Thermal point cloud. (b) Model point cloud.}
    \label{fig:R_point cloud}
\end{figure}

The generated thermal point cloud and model point cloud are shown in Figure~\ref{fig:R_point cloud}.
Thermal point clouds~(Figure.~\ref{fig:R_point cloud}~(a)) include building facades and other objects in the \gls{TIR} images, such as traffic lights, pedestrians, and vehicles.
Thermal point clouds show the geometry of building elements, including different shapes of windows.
The moldings, balconies, and special decorations are also recorded as they are.
However, the rooftop and some corners (e.g., the upper right corner) are missing due to scanning mode and height limitations.
Compared to the raw point clouds, thermal point clouds keep the original geometry features while attaching thermal attributes as intensity for the points.
The different intensities represent the temperature and can reveal inner structures like heating pipes.
The higher intensity around windows shows wooden frames and some indoor rooms with higher thermal temperatures.
Model point clouds~(Figure.~\ref{fig:R_point cloud}~(b)) generated from the \gls{LoD}3 model describe the semantic information with different colors for windows, doors, walls, roof, and ground. 
Unlike laser point clouds, where the laser penetrates the window glasses and leaves empty spaces, the model point clouds block the window areas with in-depth planes.
Moreover, all the functional segments are labeled, but non-functional decorations are simplified compared to the laser point clouds.

\begin{figure}[h!]
    \centering
    \includegraphics[width=0.8\linewidth]{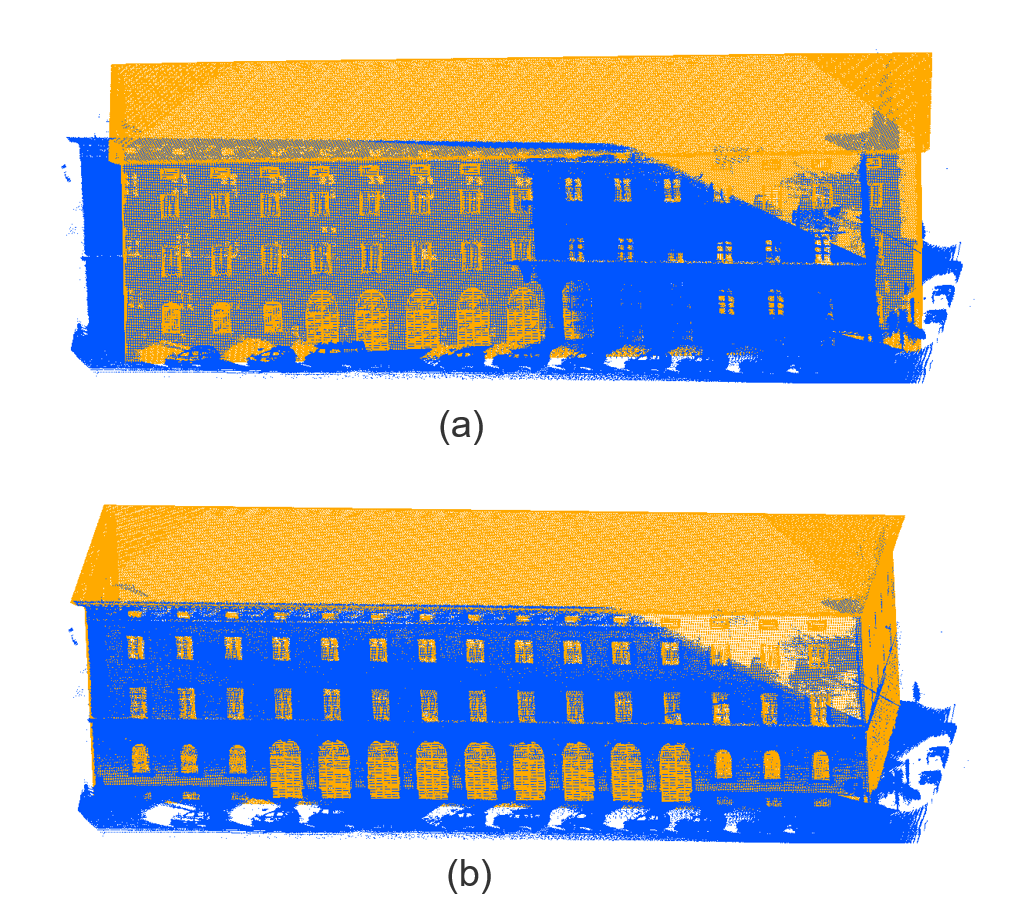}
    \caption{Registration result (a) Transformation after \gls{FGR} and (b) after fine registration. Target: orange, model point cloud; Source: blue, thermal point cloud.}
    \label{fig:R_registration}
\end{figure}

Then, we registered the thermal point clouds to the model point cloud after point cloud processing.
The results are shown in Figure.~\ref{fig:R_registration}.
Figure.~\ref{fig:R_registration}(a) illustrates the transformation of thermal point clouds with initial pose after \gls{FGR}.
The transformed thermal point cloud shows that it is roughly transformed to the model point cloud within a certain overlap.
Although the thermal point cloud has mainly limited facades with occlusions, the corresponding main planes with most of the windows were attached. 
The redundant objects on the front street and inner noisy point clouds of facades influenced the registration result.
Figure.~\ref{fig:R_registration}(b) presents the results after fine registration of the point clouds aligned with the model.
After fine registration, the thermal point clouds are correctly aligned to the facade of model point clouds. 
The corresponding windows with different shapes are equivalent, showing a satisfactory co-registration result.

%Yet, we still observe some discrepancies to the target point cloud biased towards the high-coverage density regions of the source point cloud, i.e., complete left region vs incomplete right region (blue, Figure \ref{fig:R_registration}). 
%
\begin{equation}
    fitness = \frac{\text{\# inlier correspondences}}{\text{\# points in target}}\label{eq:fitness}
\end{equation}

\begin{figure}[h!]
    \centering
    \includegraphics[width=0.8\columnwidth]{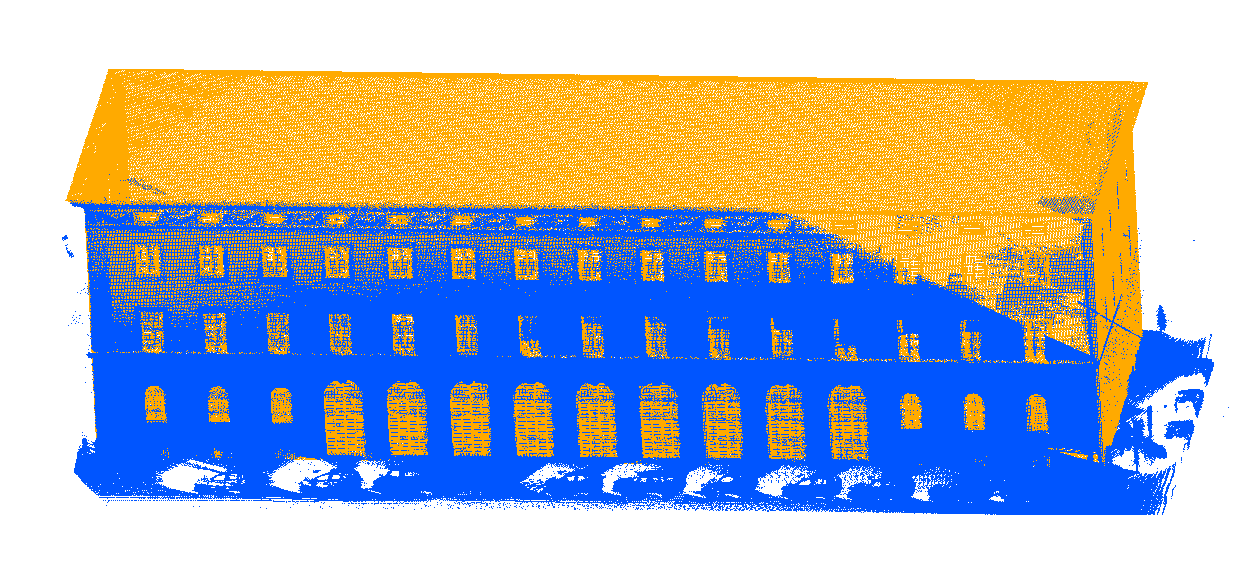}
    \caption{Manual registration result. Target: orange, model point cloud; Source: blue, thermal point cloud.}
    \label{fig:R_compare}
\end{figure}

To estimate the co-registration result, we calculate the RMSE distances~(eq.~\ref{eq:rmse}) and the fitness~(eq.\ref{eq:fitness})~(threshold = 2$m$).
As we show in Table \ref{tab:fine_registration}, our fine-registration approach can reach a high improvement rate.
In the case of our tested sample, the RMSE has decreased approximately five times (1.46m vs. 0.33m), while the fitness score has improved by approximately 65\% (0.54 vs. 0.88).
To further validate our result, a comparison experiment was conducted by manually selecting corresponding points and estimating the transformation matrix.
Six corresponding points were manually selected from model point clouds and thermal point clouds and were transformed with the estimated matrix~(Figure~\ref{fig:R_compare}).
The RMSE between the ground truth and fine registration was 0.4$m$.
Regarding the fitness and RMSE, our proposed method achieved a comparable level of accuracy and better fitness to the reference~(Table \ref{tab:fine_registration}).

\begin{figure}[h!]
    \centering
    \includegraphics[width=1.0\columnwidth]{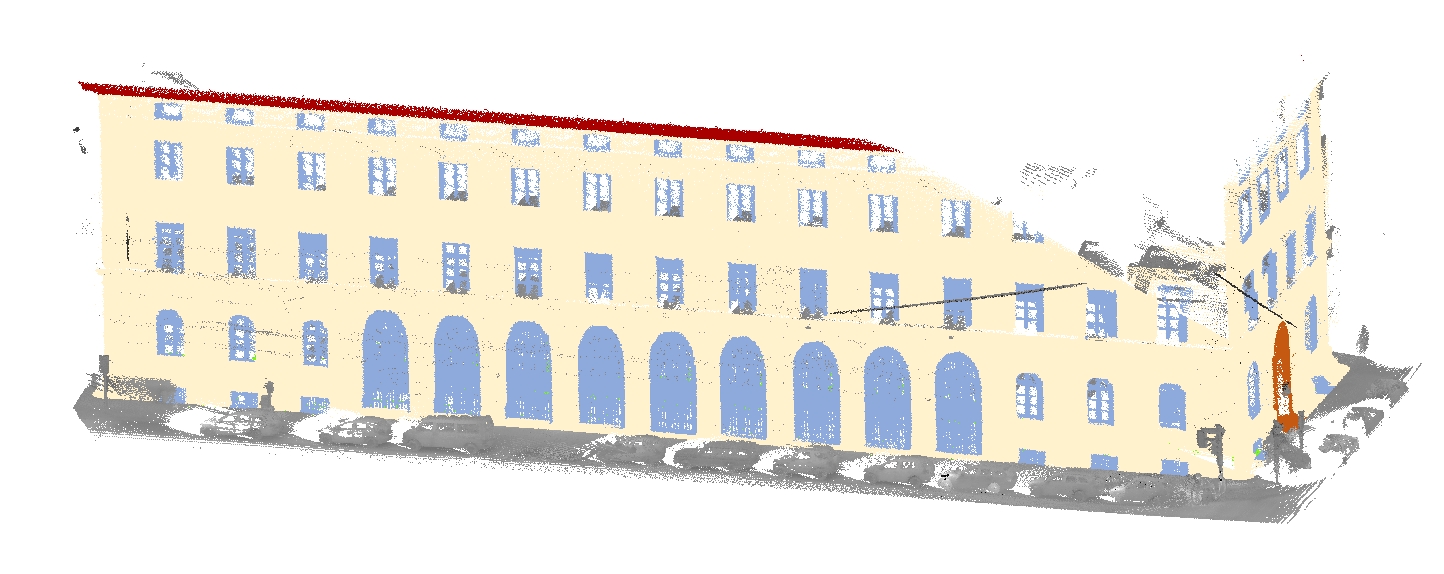}%FigX_E_pcd.png
    \caption{Semantically enriched thermal point clouds.}
    \label{fig:R_labeled}
\end{figure}

\begin{figure}[h!]
    \centering
    \includegraphics[width=1.\columnwidth]{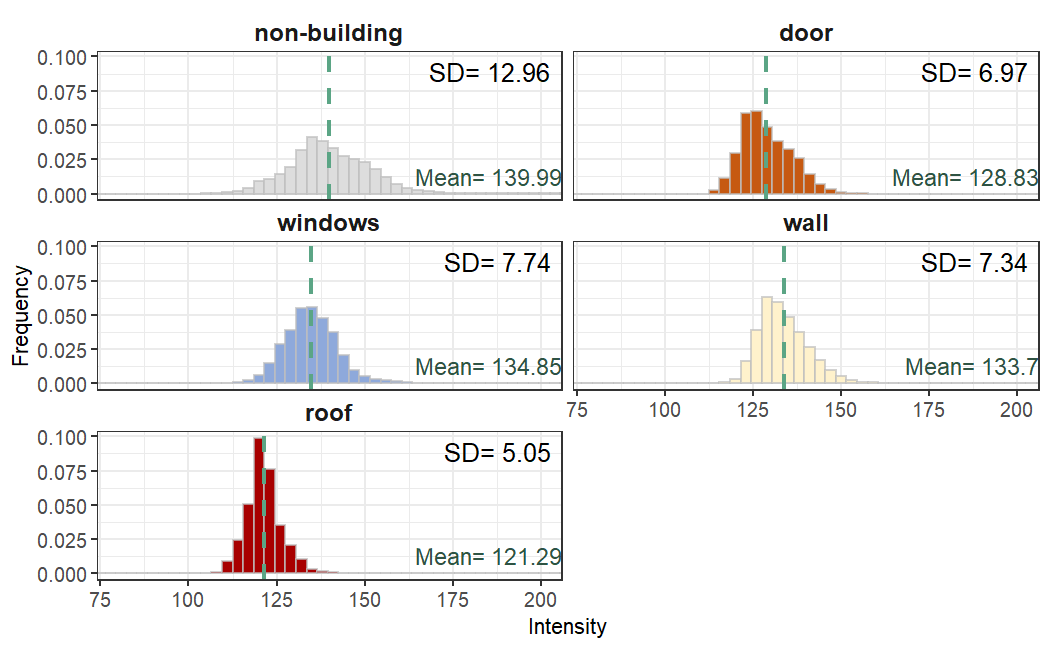}
    \caption{Semantic analysis of thermal properties and the result. The average intensity and standard deviation for each class are calculated.}
    \label{fig:R_static}
\end{figure}

With the calculated transformation matrix, the thermal point clouds can be enriched from semantic model point clouds.
By applying the transformation, the thermal point clouds are aligned to the \gls{LoD}3 model and georeferenced to the reference coordinate system w.r.t. the \gls{LoD}3 model.
The semantic labels for each laser point are calculated as described in Section.~\ref{enrichment}.
As shown in Figure.~\ref{fig:R_labeled}, the facades of the building are labeled with semantic labels while the uncovered areas~(from the street) remain unlabeled.
To further analyze the behavior of different details in the building, the statistical number is calculated based on the frequency for each class, as shown in Figure.~\ref{fig:R_static}.
For each classes, we calculate the average intensity and the standard deviation to investigate the thermal difference for different classes, which correspond to different materials and structures, in the same environment.
The average intensity value and distribution vary across the classes.
The number of wall points and windows were the prevalent classes in the point clouds.
The limited roof points below the eave show relatively lower temperature and variance compared to walls and windows.
The windows show relatively higher changes with higher standard deviation.
The initial results from this experiment show that thermal intensity varies for different materials, with potential applications such as material differentiation and anomaly detection.

%Note that the ground surface points are typically not visible from outdoor MLS point clouds during the measurement, therefore could be the noises from scanned points through windows.

\begin{table}[h!]
    \footnotesize
    \centering
    \begin{tabular}{l cc} 
	\toprule
	Method  & \multicolumn{2}{c}{vs. GT LoD3 } \\
	\cmidrule(lr){2-3}
    & Fitness $\uparrow$ & RMSE $\downarrow$ \\
    \midrule
    FGR registration & 0.54 & 1.46 \\
    %Fine registration~\cite{wysocki2021unlocking}  & {0.89} & {0.49} \\
    Fine registration~(ours)  & \textbf{0.88} & \textbf{0.33} \\
    Manually  & {0.87} & {0.33} \\
    \bottomrule
    \end{tabular}
    \caption{The co-registration results for our fine registration approach ($\uparrow$ indicates the more the better,  $\downarrow$ otherwise).}
    \label{tab:fine_registration}
\end{table}

%After semantic enrichment of the point cloud, the histogram of different classes was calculated and is shown in Figure.~\ref{fig:R_static}:

%

%As demonstrated in Figure.~\ref{fig:R_static}, the statistical number is calculated based on the frequency for each class. 

%
\section{Conclusion}\label{Conclusion}

\begin{figure}[h!]
    \centering \includegraphics[width=0.6\columnwidth]{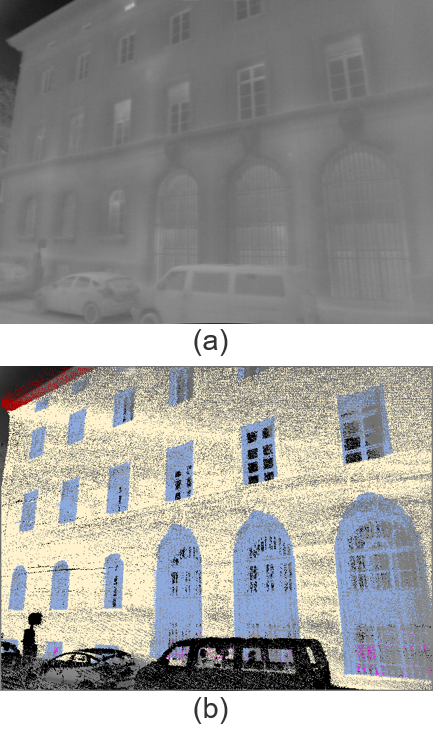}
    \caption{Semantically enriched \gls{TIR} image. (a) Original \gls{TIR} image. (b) Semantic enriched \gls{TIR} image.}
    \label{fig:C_TIR_enhance}
\end{figure}

In this paper, we propose a feasible workflow to enrich the semantic information of a thermal point cloud given a \gls{LoD}3 model.
The proposed method converts the \gls{LoD}3 model to a point cloud and registers a thermal point cloud to the model through point cloud co-registration.
With the proposed coarse-to-fine registration, the thermal point clouds can be registered to semantic model point clouds regardless of limited overlap and feature differences.
The co-registration results have comparable accuracy to the referenced manually registered results.
Finally, the semantic labels from the \gls{LoD}3 model are assigned to the thermal point clouds for analysis.
This work is not limited to thermal point clouds but also to all the co-registration tasks, from laser point clouds to building models requiring semantic label transfer.
The enriched results improve the point cloud labeling pace by giving knowledge and enhancing the efficiency of semantic data generation.
The thermal point clouds with \gls{LoD}3 labels can serve as supportive data for further urban study and algorithm development, such as testing and training deep learning models.
With the given pose of the image, the labels can be back-projected to the \gls{TIR} images for processing and supportive analysis, as shown in Figure~\ref{fig:C_TIR_enhance}.
In this work, we combine point cloud geometry for thermal anomaly interpretation, but also a bi-directional information exchange can be pursued: The thermal properties might be mapped onto the LoD3 objects enriching the LoD3 in radiometric thermal features for visualization and building operation monitoring~\cite{manoj}
%Despite combing point cloud geometry for thermal anomaly interpretation, a bi-directional information exchange that maps the thermal properties onto the LoD3 objects can enrich the LoD3 in radiometric thermal features for visualization and building operation monitoring~\cite{manoj}.

For future work, it is worth further investigation into the topic of robust methods for model and point cloud co-registration, especially in large-scale datasets.
Though this work proposed the initial tasks for single building co-registration and semantic enrichment results, how to improve the efficiency is to be explored.
How to use the features in the laser point clouds and \gls{LoD}3 model while minimizing the effect introduced by different feature representations is still a problem.
The co-registration results can help enriching and fusing information from different datasets or localize and compare the changes from other models or time stamps.

 \section*{Acknowledgement}

% KAO: Sloppy spacing ensures non-overfull lines. Can be removed if this is not an issue.
We would like to express our gratitude to the Leonhard Obermeyer Center (LOC) at the Technical University of Munich (TUM) for their support in facilitating this work. 
Additionally, we extend our appreciation to tum2twin team at TUM for their valuable insights and for generously providing access to the CityGML datasets.
\sloppy

% KAO: Non-breaking space
%\section{References}\label{sec:References}

{
	\begin{spacing}{1.17}
		\normalsize
		\bibliography{3DGeoInfo} % Include your own bibliography (*.bib), style is given in isprs.cls

\begin{thebibliography}{xx}

\bibitem[Abreu et al., 2023]{Abreu2023}
Abreu, N., Pinto, A., Matos, A., Pires, M., 2023.
 Procedural point cloud modelling in scan-to-BIM and scan-vs-BIM applications: a review.
 {\em ISPRS International Journal of Geo-Information}, 12(7), 260.

\bibitem[Aiger et al., 2008]{aiger20084}
Aiger, D., Mitra, N.~J., Cohen-Or, D., 2008.
 4-points congruent sets for robust pairwise surface registration.
 \emph{ACM SIGGRAPH 2008 Papers}, SIGGRAPH '08, Association for Computing Machinery, New York, NY, USA.

\bibitem[Barnea and Filin, 2008]{barnea2008keypoint}
Barnea, S., Filin, S., 2008.
 Keypoint based autonomous registration of terrestrial laser point-clouds.
 {\em ISPRS Journal of Photogrammetry and Remote Sensing}, 63(1), 19--35.

\bibitem[Biljecki et al., 2015]{biljeckiApplications3DCity2015}
Biljecki, F., Stoter, J., Ledoux, H., Zlatanova, S., {\c C}{\"o}ltekin, A., 2015.
 Applications of {3D} city models: {State} of the art review.
 {\em ISPRS International Journal of Geo-Information}, 4(4), 2842--2889.

\bibitem[Biswanath et al., 2023]{manoj}
Biswanath, M.~K., Hoegner, L., Stilla, U., 2023.
 Thermal Mapping from Point Clouds to 3D Building Model Facades.
 {\em Remote Sensing}, 15(19).

\bibitem[Bosch{\'e}, 2012]{Bosche2012}
Bosch{\'e}, F., 2012.
 Plane-based registration of construction laser scans with 3D/4D building models.
 {\em Advanced Engineering Informatics}, 26(1), 90--102.

\bibitem[Chen et al., 2022]{Chen2022}
Chen, J., Li, S., Lu, W., 2022.
 Align to locate: Registering photogrammetric point clouds to BIM for robust indoor localization.
 {\em Building and Environment}, 209, 108675.

\bibitem[Chen and Yu, 2019]{chen2019feature}
Chen, X., Yu, K., 2019.
 Feature line generation and regularization from point clouds.
 {\em IEEE Transactions on Geoscience and Remote Sensing}, 57(12), 9779--9790.

\bibitem[Goebbels and Pohle-Fr{\"o}hlich, 2018]{goebbels2018line}
Goebbels, S., Pohle-Fr{\"o}hlich, R., 2018.
 Line-based registration of photogrammetric point clouds with 3{D} city models by means of mixed integer linear programming.
 \emph{VISIGRAPP (4: VISAPP)}, 299--306.

\bibitem[Goebbels et al., 2019]{goebbels2019iterative}
Goebbels, S., Pohle-Fr\"ohlich, R., Pricken, P., 2019.
 Iterative Closest Point algorithm for accurate registration of coarsely registered point clouds with CityGML models.
 {\em ISPRS Annals of the Photogrammetry, Remote Sensing and Spatial Information Sciences}, IV-2/W5, 201--208.

\bibitem[Gr\"oger et al., 2012]{grogerOGCCityGeography2012}
Gr\"oger, G., Kolbe, T.~H., Nagel, C., H\"afele, K.-H., 2012.
 {{OGC City Geography Markup Language CityGML Encoding Standard version 2.0}}.
 Open Geospatial Consortium: Wayland, MA, USA, 2012.

\bibitem[Gruner et al., 2022]{Gruner2022}
Gruner, F., Romanschek, E., Wujanz, D., Clemen, C., 2022.
 Co-registration of TLS point clouds with scan-patches and BIM-faces.
 {\em The International Archives of the Photogrammetry, Remote Sensing and Spatial Information Sciences}, XLVI-5/W1-2022, 109--114.

\bibitem[Haala and Kada, 2010]{HAALA2010570}
Haala, N., Kada, M., 2010.
 An update on automatic {3D} building reconstruction.
 {\em ISPRS Journal of Photogrammetry and Remote Sensing}, 65(6), 570 - 580.

\bibitem[Hoegner and Gleixner, 2022]{hoegner2022automatic}
Hoegner, L., Gleixner, G., 2022.
 Automatic extraction of facades and windows from MLS point clouds using VoxelSpace and visibility analysis.
 {\em The International Archives of the Photogrammetry, Remote Sensing and Spatial Information Sciences}, XLIII-B2-2022, 387--394.

\bibitem[Huang et al., 2020]{helmutMayerLoD3}
Huang, H., Michelini, M., Schmitz, M., Roth, L., Mayer, H., 2020.
 {LoD3} Building Reconstruction from Multi-source images.
 {\em The International Archives of the Photogrammetry, Remote Sensing and Spatial Information Sciences}, XLIII-B2-2020, 427--434.

\bibitem[Janßen et al., 2022]{JanßenKuhlmannHolst+2022+91+106}
Janßen, J., Kuhlmann, H., Holst, C., 2022.
 Target-based Terrestrial Laser Scan Registration extended by Target Orientation.
 {\em Journal of Applied Geodesy}, 16(2), 91--106.

\bibitem[Janßen et al., 2023]{JanßenKuhlmannHolst+2023}
Janßen, J., Kuhlmann, H., Holst, C., 2023.
 Keypoint-based registration of {TLS} point clouds using a statistical matching approach.
 {\em Journal of Applied Geodesy}.

\bibitem[Kaiser et al., 2022]{Kaiser2022}
Kaiser, T., Clemen, C., Maas, H.-G., 2022.
 Automatic co-registration of photogrammetric point clouds with digital building models.
 {\em Automation in construction}, 134, 104098.

\bibitem[Kolbe, 2009]{Kolbe2009}
Kolbe, T.~H., 2009.
 {\em Representing and exchanging 3{D} city models with {CityGML}}.
 Springer, Berlin, Heidelberg.

\bibitem[Kolbe and Donaubauer, 2021]{Kolbe2021}
Kolbe, T.~H., Donaubauer, A., 2021.
 {\em Semantic 3D City Modeling and {BIM}. In:Urban Informatics}.
 Springer, Singapore.

\bibitem[Kolbe et al., 2021]{kolbeOGCCityGeography2021}
Kolbe, T.~H., Kutzner, T., Smyth, C.~S., Nagel, C., Roensdorf, C., Heazel, C., 2021.
 {{OGC City Geography Markup Language}} ({{CityGML}}) {{Part}} 1: {{Conceptual Model Standard}} v3.0, {O}pen geospatial consortium.

\bibitem[Kutzner et al., 2020]{Kutzner2020}
Kutzner, T., Chaturvedi, K., Kolbe, T.~H., 2020.
 CityGML 3.0: New Functions Open Up New Applications.
 {\em PFG – Journal of Photogrammetry, Remote Sensing and Geoinformation Science}, 88(1), 43--61.

\bibitem[Ledoux et al., 2019]{ledoux2019cityjson}
Ledoux, H., Arroyo~Ohori, K., Kumar, K., Dukai, B., Labetski, A., Vitalis, S., 2019.
 CityJSON: A compact and easy-to-use encoding of the CityGML data model.
 {\em Open Geospatial Data, Software and Standards}, 4(1), 1--12.

\bibitem[Li et al., 2022]{li2022point}
Li, J., Zhan, J., Zhou, T., Bento, V.~A., Wang, Q., 2022.
 Point cloud registration and localization based on voxel plane features.
 {\em ISPRS Journal of Photogrammetry and Remote Sensing}, 188, 363--379.

\bibitem[Lu et al., 2019]{lu2019deepvcp}
Lu, W., Wan, G., Zhou, Y., Fu, X., Yuan, P., Song, S., 2019.
 Deepvcp: An end-to-end deep neural network for point cloud registration.
 \emph{Proceedings of the IEEE/CVF international conference on computer vision}, 12--21.

\bibitem[Lucks et al., 2021]{lucks2021improving}
Lucks, L., Klingbeil, L., Pl{\"u}mer, L., Dehbi, Y., 2021.
 Improving trajectory estimation using {3D} city models and kinematic point clouds.
 {\em Transactions in GIS}, 25(1), 238--260.

\bibitem[Matrone et al., 2020]{matrone2020comparing}
Matrone, F., Grilli, E., Martini, M., Paolanti, M., Pierdicca, R., Remondino, F., 2020.
 Comparing machine and deep learning methods for large {3D} heritage semantic segmentation.
 {\em ISPRS International Journal of Geo-Information}, 9(9), 535.

\bibitem[Mellado et al., 2014]{mellado2014super}
Mellado, N., Aiger, D., Mitra, N.~J., 2014.
 Super 4pcs fast global pointcloud registration via smart indexing.
 \emph{Computer graphics forum},  33-5, Wiley Online Library, 205--215.

\bibitem[Ramón et al., 2022]{ramonReview}
Ramón, A., Adán, A., {Javier Castilla}, F., 2022.
 Thermal point clouds of buildings: A review.
 {\em Energy and Buildings}, 274, 112425.

\bibitem[Roschlaub and Batscheider, 2016]{RoschlaubBatscheider}
Roschlaub, R., Batscheider, J., 2016.
 An {INSPIRE}-conform {3D} building model of {Bavaria} using cadastre information, {LiDAR} and image matching.
 {\em The International Archives of the Photogrammetry, Remote Sensing and Spatial Information Sciences}, XLI-B4, 747--754.

\bibitem[Rusinkiewicz and Levoy, 2001]{Rusinkiewicz2001}
Rusinkiewicz, S., Levoy, M., 2001.
 Efficient variants of the {ICP} algorithm.
 \emph{Proceedings Third International Conference on 3-D Digital Imaging and Modeling}, 145--152.

\bibitem[Rusu and Cousins, 2011]{rusu20113d}
Rusu, R.~B., Cousins, S., 2011.
 3d is here: Point cloud library (pcl).
 \emph{2011 IEEE international conference on robotics and automation}, IEEE, 1--4.

\bibitem[Schnabel et al., 2007]{schnabel2007efficient}
Schnabel, R., Wahl, R., Klein, R., 2007.
 Efficient {RANSAC} for point-cloud shape detection.
 {\em Computer Graphics Forum}, 26(2), 214-226.

\bibitem[Segal et al., 2009]{segal2009generalized}
Segal, A., Haehnel, D., Thrun, S., 2009.
 Generalized-{ICP}.
 \emph{Robotics: Science and Systems},  2-4, Seattle, WA, 435.

\bibitem[Sheik et al., 2022]{Sheik2022}
Sheik, N.~A., Deruyter, G., Veelaert, P., 2022.
 Plane-based robust registration of a building scan with its BIM.
 {\em Remote Sensing}, 14(9), 1979.

\bibitem[Su et al., 2022]{SU2022108372}
Su, Y., Liu, W., Yuan, Z., Cheng, M., Zhang, Z., Shen, X., Wang, C., 2022.
 DLA-Net: Learning dual local attention features for semantic segmentation of large-scale building facade point clouds.
 {\em Pattern Recognition}, 123, 108372.

\bibitem[Wysocki et al., 2022]{wysockiUnderpasses}
Wysocki, O., Hoegner, L., Stilla, U., 2022.
 {Refinement of semantic 3D building models by reconstructing underpasses from MLS point clouds}.
 {\em International Journal of Applied Earth Observation and Geoinformation}, 111, 102841.

\bibitem[Wysocki et al., 2023a]{wysockiMLS2LoD3}
Wysocki, O., Hoegner, L., Stilla, U., 2023a.
 MLS2LoD3: Refining low LoDs building models with MLS point clouds to reconstruct semantic LoD3 building models.
 {\em Accepted to proceedings of 3D GeoInfo 2023, Lecture Notes in Geoinformation and Cartography}.

\bibitem[Wysocki et al., 2023b]{wysocki2023scan2lod3}
Wysocki, O., Xia, Y., Wysocki, M., Grilli, E., Hoegner, L., Cremers, D., Stilla, U., 2023b.
 {Scan2LoD3}: Reconstructing semantic {3D} building models at {LoD3} using ray casting and {Bayesian} networks.
 {\em IEEE/CVF Conference on Computer Vision and Pattern Recognition Workshops (CVPRW)}, 6547--6557.

\bibitem[Wysocki et al., 2021]{wysocki2021unlocking}
Wysocki, O., Xu, Y., Stilla, U., 2021.
 Unlocking point cloud potential: {Fusing} {MLS} point clouds with semantic {3D} building models while considering uncertainty.
 {\em {ISPRS} - Int. Ann. Photogramm. Remote Sens. Spatial Inf. Sci.}, VIII-4/W2, 45--52.

\bibitem[Xu et al., 2023]{xu2023point}
Xu, N., Qin, R., Song, S., 2023.
 Point cloud registration for {L}iDAR and photogrammetric data: A critical synthesis and performance analysis on classic and deep learning algorithms.
 {\em ISPRS open journal of Photogrammetry and Remote Sensing}, 8, 100032.

\bibitem[Yang et al., 2015]{yang2015go}
Yang, J., Li, H., Campbell, D., Jia, Y., 2015.
 Go-{ICP}: A globally optimal solution to 3D ICP point-set registration.
 {\em IEEE Transactions on Pattern Analysis and Machine Intelligence}, 38(11), 2241--2254.

\bibitem[Zhang et al., 2022]{rs14122883}
Zhang, R., Li, G., Wiedemann, W., Holst, C., 2022.
 KdO-Net: Towards Improving the Efficiency of Deep Convolutional Neural Networks Applied in the 3D Pairwise Point Feature Matching.
 {\em Remote Sensing}, 14(12).

\bibitem[Zhou et al., 2016]{zhou2016fast}
Zhou, Q., Park, J., Koltun, V., 2016.
 Fast global registration.
 \emph{Computer Vision--ECCV 2016: 14th European Conference, Amsterdam, The Netherlands, October 11-14, 2016, Proceedings, Part II 14}, Springer, 766--782.

\bibitem[Zhou et al., 2018]{zhou2018open3d}
Zhou, Q., Park, J., Koltun, V., 2018.
 Open3D: A modern library for 3{D} data processing.
 {\em arXiv preprint arXiv:1801.09847}.

\bibitem[Zhu et al., 2020]{zhu2020}
Zhu, J., Gehrung, J., Huang, R., Borgmann, B., Sun, Z., Hoegner, L., Hebel, M., Xu, Y., Stilla, U., 2020.
 TUM-MLS-2016: An annotated mobile LiDAR dataset of the TUM city campus for semantic point cloud interpretation in urban areas.
 {\em Remote Sensing}, 12(11), 1875.

\bibitem[Zhu et al., 2023]{zhu2023generation}
Zhu, J., Xu, Y., Hoegner, L., Stilla, U., 2023.
 Generation of Thermal Point Clouds From Uncalibrated Thermal Infrared Image Sequences and Mobile Laser Scans.
 {\em IEEE Transactions on Instrumentation and Measurement}, 72, 1-16.

\bibitem[Zhu et al., 2021]{zhu2021fusion}
Zhu, J., Xu, Y., Ye, Z., Hoegner, L., Stilla, U., 2021.
 Fusion of urban 3D point clouds with thermal attributes using MLS data and TIR image sequences.
 {\em Infrared Physics \& Technology}, 113, 103622.

\end{thebibliography}
	\end{spacing}
}

\end{document}